\documentclass[runningheads]{llncs}
\usepackage{graphicx}

\usepackage{subcaption}
\usepackage{tikz}
\usepackage{comment}
\usepackage{amsmath,amssymb} 
\usepackage{color}
\usepackage{stmaryrd}
\usepackage{hyperref}
\usepackage[accsupp]{axessibility}  

\begin{document}
\pagestyle{headings}
\mainmatter
\def\ECCVSubNumber{11}  

\title{Privacy-Preserving Person Detection Using Low-Resolution Infrared Cameras}

\titlerunning{Privacy-Preserving Person Detection Using Low-Resolution Infrared Cameras}
%
\author{Thomas Dubail \and
Fidel Alejandro Guerrero Peña\index{Guerrero Peña,Fidel Alejandro} \and
Heitor Rapela Medeiros\index{Medeiros,Heitor Rapela} \and
Masih Aminbeidokhti \and
Eric Granger \and
Marco Pedersoli}
\authorrunning{T. Dubail et al.}
%
\institute{Laboratoire d'imagerie, de vision et d'intelligence artificielle (LIVIA)\\ Dept. of Systems Engineering,  ETS Montreal, Canada\\
\email{ \{thomas.dubail.1, fidel-alejandro.guerrero-pena.1, heitor.rapela-medeiros.1, masih.aminbeidokhti.1\}@ens.etsmtl.ca,\\ \{eric.granger, marco.pedersoli\}@etsmtl.ca} }
\maketitle

\begin{abstract}
In intelligent building management, knowing the number of people and their location in a room are important for better control of its illumination, ventilation, and heating with reduced costs and improved comfort. This is typically achieved by detecting people using compact embedded devices that are installed on the room's ceiling, and that integrate low-resolution infrared camera, which conceals each person's identity. However, for accurate detection, state-of-the-art deep learning models still require supervised training using a large annotated dataset of images. In this paper, we investigate cost-effective methods that are suitable for person detection based on low-resolution infrared images. Results indicate that for such images, we can reduce the amount of supervision and computation, while still achieving a high level of detection accuracy. Going from single-shot detectors that require bounding box annotations of each person in an image, to auto-encoders that only rely on unlabelled images that do not contain people, allows for considerable savings in terms of annotation costs, and for models with lower computational costs. We validate these experimental findings on two challenging top-view datasets with low-resolution infrared images.

\end{abstract}

\section{Introduction}

Intelligent building management solutions seek to maximize the comfort of occupants, while minimizing energy consumption. These types of solutions are crucial for reducing the use of fossil fuels with a direct impact on the environment. Such energy-saving is usually performed by adaptively controlling lighting, heating, ventilation, and air-conditioning (HVAC) systems based on building occupancy, and in particular the number of people present in a given room. For this, low-cost methods are needed to assess the level of room occupancy, and efficiently control the different systems within the building.

Among the different levels of occupancy information that can be extracted in an intelligent building \cite{sun2020review}, Sun et al. define the location of occupants as the most important and fine-grained for smart building control. Given the recent advances in machine learning and computer vision, most solutions usually rely on deep convolutional neural networks (CNNs) to detect people \cite{gao2016people,chen2018unobtrusive}. Despite the high level of accuracy that can be achieved with CNNs for visual object detection based on RGB images, their implementation for real-world video surveillance applications incurs in a high computational complexity, privacy issues, and gender and race biases \cite{buolamwini2018gender,schwemmer2020diagnosing}. Finally, building occupancy management solutions are typically implemented on compact embedded devices, rigidly installed on the ceiling or portals of rooms, and integrating inexpensive cameras that can capture low-resolution IR images.

To mitigate these issues, He et al.~\cite{he2021privacy} have proposed a privacy-preserving object detector that blurs people's faces before performing detection. To strengthen the detector against gender/race biases, the same authors proposed a face-swapping variation that also preserves privacy at the cost of increased computational complexity. Regardless of the good performance, their approach does not ensure confidentiality at the acquisition level, relying on RGB sensors to build the solution. Furthermore, their detector was designed for fully annotated settings using COCO \cite{lin2014microsoft} as the base dataset. This makes it difficult to generalize to people detection under different capture conditions (like when cameras are located on the ceiling on compact embedded systems), and extreme changes in the environment. In addition, it is difficult to collect and annotate image data to train or fine-tune CNN-based object detectors for a given application, so weakly-supervised or unsupervised training is a promising approach.

In contrast, our work tackles the occupants location problem by detecting people in infrared (IR) images at low resolution, which avoids most of the above-mentioned issues on privacy. Low resolution not only reduces computational complexity but also improves privacy, i.e., a detection on high-resolution infrared images would not be enough as it is possible to re-identify people \cite{electronics11030454}. More specifically, we analyse people detection with different levels of supervision. In this work, we compare unsupervised, weakly-supervised and fully supervised solutions. This is an essential aspect of the detection pipeline since producing bounding box annotations is very expensive, and there is a lack of good open-source object detection datasets for low-resolution infrared scenarios. In fact, reducing the level of supervision can lead to improved scalability for real applications and reduced computation, which is important considering the use of the proposed algorithms on embedded devices.

The contributions of this work are the following. (i) We propose cost-effective methods for estimating room occupancy under a low-supervision regime based on low-resolution IR images, while preserving users privacy. (ii) We provide an extensive empirical comparison of several cost-effective methods that are suitable for person detection using low-resolution IR cameras. Results indicate that, using top-view low-resolution images, methods that rely on weakly-labeled image data can provide good detection results, and thus save annotation efforts and reduce the required complexity of the detection model. (iii) To investigate the performance of person detecting methods on low-resolution IR images, our results are shown in two challenging datasets -- the FIR-Image-Action and Distech-IR datasets. Finally, we provide bounding box annotations for the FIR-Image-Action \cite{FIR-Image-Action-Dataset} dataset\footnote{\url{https://github.com/ThomasDubail/FIR-Image-Action-Localisation-Dataset}}.

\section{Related Work}

Privacy-preserving methods are of great interest to the scientific community. As a result, multiple approaches have been proposed in the past to circumvent the challenge. Ryoo et al.~\cite{ryoo2017privacy} proposed a method for learning a transformation that obtained multiple low-resolution images from a high-resolution RGB source. The method proved effective for action recognition even when inputs' resolution were down-sampled to $16\times 12$. These findings were validated by others~\cite{wang2016studying,chen2017semi} using similar downsampling-based techniques to anonymize the people displayed in the images. On the other hand, recent approaches~\cite{ren2018learning,he2021privacy} have focused on producing blurred or artificial versions of people's faces while preserving the rest of the image intact. These methods usually rely on Generative Adversarial Networks (GANs) to preserve image utility, while producing unidentifiable faces. Specifically, the method of He et al.~\cite{he2021privacy} is one of the first approaches to apply such anonymization in an object detection task. Despite recent advances in the area, these proposals are specialized for RGB images, which are anonymized after the undisclosed acquisition. As an alternative to RGB cameras, others authors~\cite{taoHomeActivityMonitoring2018,lilitao3DConvolutionalNeural2019,tatenoPrivacyPreservedFallDetection2020,tatenoHumanMotionDetection2020} have proposed using low-resolution IR cameras to preserve anonymity at the phase of image acquisition. While most of these works target action recognition task, we focus on people detection as in~\cite{shengshengyuRobustMethodDetecting2008,caoCountingPeopleUsing2016}.

Complementary to privacy preservation, this work also focuses on studying detection methods with different levels of supervision. Such a study aims to find techniques that reduce annotation costs without a significant performance reduction when compared with fully supervised approaches. In this work, we followed the auto-encoders (AE)-based anomaly-detection method similar to the one proposed by Baur et al.~\cite{baurDeepAutoencodingModels2019}. The technique is used to learn the distribution of typical cases, and applied later to identify abnormal regions within the image. However, different from their proposal, we focus on object detection instead of image segmentation. We also evaluate weakly-supervised detection methods based on Class Activation Maps (CAMs)~\cite{zhouLearningDeepFeatures2015,selvarajuGradCAMVisualExplanations2020} following the authors algorithm. Nonetheless, we use a customized CNN to be consistent with our low-resolution inputs. Finally, we consider also the fully supervised techniques Single Shot Detector~\cite{liuSSDSingleShot2016} and Yolo v5~\cite{glenn_jocher_2022_6222936} as upper-bound references, sharing the same backbone as the previously described approaches. The aim of this work is not to use the latest developments for each type of technique. Instead, we aim to achieve a good trade-off between performance and computational complexity with simple and commonly used techniques. In particular, we favor simple occupants' detection approaches that can run in embedded devices and with capabilities for handling low-resolution infrared images.

\section{Person Detection with Different Levels of Supervision}

Several cost-effective methods may be suitable for person detection using low-resolution IR cameras installed on ceilings. The goal of object detection is to find a mapping $f_\theta$ such that $f_\theta(x)=z$, where $z$ are the probabilities that a bounding box belongs to each class. Note that such a mapping can be obtained using any level of supervision.  In this paper, we seek to compare the detection accuracy of methods that rely on different levels of image annotation, and thereby assess the complexity needed to design embedded person detection systems.  

In this work we consider a ``fully annotated" dataset of IR images at low resolution $\mathcal{F}=\{(x_0,b_0),(x_1,b_1),...,(x_N,b_N)\}$, with $x$ an IR image and $b=\{(c_0,d_0,w_0,h_0),...,(c_B,d_B,w_B,h_B)\}$ rectangular regions enclosing the objects of interest, also known as bounding boxes. Without loss of generality, we use a center pixel representation \emph{(center x, center y, width, height)} \cite{liuSSDSingleShot2016} for defining the bounding boxes. In the given formulation, all bounding boxes belong to the persons' category.
Consequently, a ``weakly annotated" IR dataset is defined as $\mathcal{W}=\{(x_0,y_0),...,$
$(x_N,y_N)\}$ in which $y_i\in \{0,1\}$ corresponds to an image-level annotation indicating whether a person is present ($y_i=1$) or not ($y_i=0$) in the image $x_i$. Finally, at the lowest level of supervision, an "unlabeled" dataset containing only IR images without annotations is expressed as $\mathcal{U}=\{x_i\}$. Please note that for this study, the datasets $\mathcal{F}$, $\mathcal{W}$, and $\mathcal{U}$ are drawn from the same pool of IR images but with different levels of annotations.

The rest of this section details the methods compared in this paper, each one trained according to a different level of supervision. Here, the backbone of all deep learning based methods remained the same.  We focus on low-cost methods that can potentially be implemented on compact embedded devices.   

\subsection{Detection through thresholding}

\begin{figure}[ht]
\centering
\begin{tabular}{ccc}
\includegraphics[width=.35\linewidth]{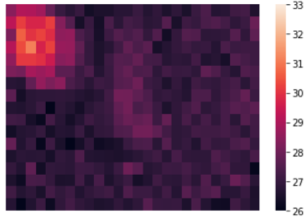} &
\includegraphics[width=.3\linewidth]{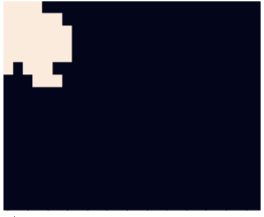} &
\includegraphics[width=.3\linewidth]{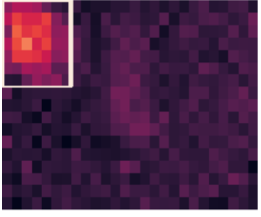} \\
(a) & (b) & (c) \\
\end{tabular}
\caption{Example of an IR image (a) that is binarized using the threshold $\tau=29$ (b), and represented as a bounding box (c).}
\label{fig:threshold}
\end{figure}

Let $x$ be an IR thermal image from $\mathcal{U}$. The people within the images appear as high-temperature blobs easily distinguishable from the low-temperature background, \autoref{fig:threshold}a. Such a property allows us to directly apply a threshold-based mapping $g_{\tau}(x) = \Phi(\llbracket x\geq\tau \rrbracket)$ to obtain the persons' location. In this formulation, we use $\llbracket\cdot \rrbracket$ to refer to the Iverson bracket notation, 
which denotes the binarization of the image $x$ according to the threshold $\tau$, \autoref{fig:threshold}b. Here, the value of $\tau$ is manually determined according to a validation set, 
or automatically following Otsu's method \cite{4310076}, hereafter referred as \emph{Threshold} and \emph{Otsu's Threshold} respectively. Finally, a mapping $\Phi$ converts the segmentation map into a bounding box by taking the minimum and maximum pixels from each binary blob (see \autoref{fig:threshold}c).
This method is more seen as a post-processing step for the following methods, although we have evaluated it to have a lower bound for detection.

\subsection{Unsupervised anomaly-based detection using auto-encoders}

\begin{figure}[!b]
\centering
\includegraphics[width=12cm]{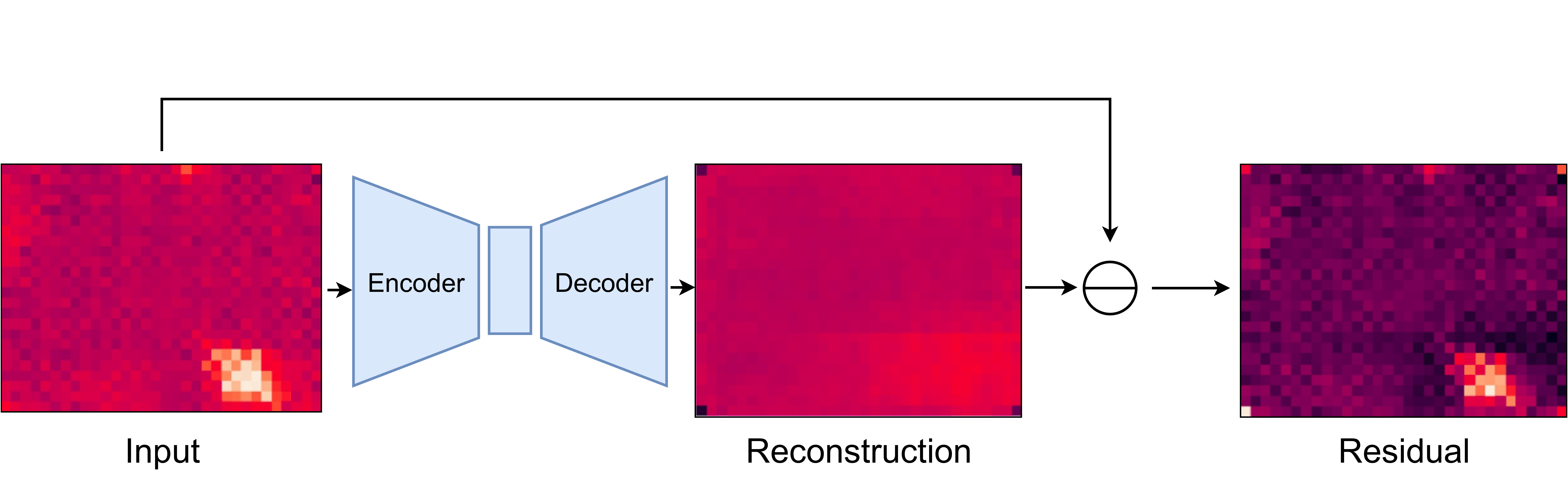}
\caption{Unsupervised anomaly-based method for person detection from low-resolution IR images.}
\label{fig:auto_encoder}
\end{figure}

For this next approach, people are considered anomalies for the distribution of empty rooms. In this work, we follow the method proposed by Baur et al. \cite{baurDeepAutoencodingModels2019} to model the background distribution using an auto-encoder (AE) $f_\theta$ trained using only empty rooms, $\mathcal{W}^0=\{x_i\mid y_i=0\}$. Such an approach acts as a background reconstruction technique whenever an anomaly is present, i.e., the AE will not be able to reconstruct it. Then, we can highlight the anomaly by taking the difference between the input image and the obtained reconstruction, $x-f_\theta(x)$ (see \autoref{fig:auto_encoder}). Finally, the anomaly detection method for person detection is defined as $\Lambda_{\theta,\tau}(x)= g_\tau(x-f_\theta (x))$ where $g_\tau$ is the thresholding technique explained in the previous section. Thus, the detection is performed in a two-step process: anomaly boosting and anomaly segmentation-localization, which can be done by setting a threshold $\tau$.

The encoder architecture comprises six convolutional layers with kernels of size $3\times 3$. Max-pooling operations are used every two convolutional layers to increase the field of view. The decoder follows a symmetrical architecture of the encoder using transposed convolution with a stride of 2 as upsampling technique. The bottleneck uses a linear layer with 256 neurons which encode input information as a vector projected onto the latent space. Finally, a reconstruction loss is used to guide the training process and find a feasible minimizer $\theta$ for solving our background reconstruction task. In this work the Mean Square Error (MSE) loss is used, $\mathcal{L}_{MSE}(x, \theta) = \displaystyle\frac{1}{|W^0|} \sum_{x_i\in W^0} ( x_i - f_\theta(x_i) )^2$. In later sections we refer to this approach as \emph{deep auto-encoder} or simply \emph{dAE}.

Besides the classical AE, several types of hourglass architectures have been used for anomaly detection \cite{baurDeepAutoencodingModels2019}. One of the most popular versions are the \emph{variational auto-encoders} (\emph{dVAE})~\cite{kingmaAutoEncodingVariationalBayes2014} where the latent vector is considered to be drawn from a given probabilistic distribution. Here we use the same architecture for both AE-based methods with a distinction for the loss function where the KL-divergence regularization is added to the MSE loss to enforce a normal distribution to the latent space.

\subsection{Weakly-supervised detection using class activation mapping}

Let $(x,y)$ be a generic tuple from $\mathcal{W}$ where $x$ is an image and $y\in\{0,1\}$ is a category indicating whether a person is present or not in $x$. A weakly-supervised approach for object localization is such that, by exploiting only the image-level annotation during training, learns a mapping $c_\varphi$ to retrieve the object location during the evaluation. This kind of approaches based on Class Activation Maps (CAM) techniques \cite{yangCombinationalClassActivation2020,kitanoClassificationLocalizationDisease,baeRethinkingClassActivation2020} has been widely explored in the literature. The principle is based on the use of the compound function $c_\varphi(x)=(c_\psi^1\circ c_\phi^0)(x)$ that relies on a feature extractor $c_\phi^0$ followed by a binary classifier $c_\psi^1$. Here $c_\phi^0$ is implemented as a CNN and $c_\psi^1$ as a Multi-layer Perceptron. Then, the following minimization based on cross-entropy loss is performed in order to find the optimal set of weights: $\min_\theta -\sum_{(x,y)\in \mathcal{W}} y\cdot \log c_\varphi(x)$. 
Once a feasible set of weights is found, a non-parametric transformation function uses the output from the feature extractor $c_\phi^0$ to produce an activation map for each category in the task, $M(c_\phi^0(x))$. Note that in this task, the computation of such an activation map is only performed whenever a positive classification is obtained, i.e., $c_\varphi(x)\geq 0.5$. The architecture used for $c_\phi^0$ is the same as in the encoder for the \emph{dAE} technique, but without using the Max-pooling layers to avoid losing resolution. In this work, we use three variants for the Global Average Pooling $M$. The first is the classic weighting approach proposed by \cite{zhouLearningDeepFeatures2015}, hereafter referred to as \emph{CAM}, the second is the gradient-based CAM proposed by \cite{selvarajuGradCAMVisualExplanations2020}, known as \emph{GradCAM}, and the last one is the hierarchical approach known as \emph{LayerCAM}~\cite{jiang2021layercam}. As in the previous techniques, the final localization is obtained using the thresholding-based mapping $g_\tau$.

\subsection{Fully-supervised detection using single shot detectors}

At the higher level of supervision, we explore the mapping function with bounding box annotations within the cost function. Let $h_\vartheta$ be a mapping parameterized over $\vartheta$, which produces bounding box predictions. Among the different types of detectors existing in the literature, we used Single Shot Detector (\emph{SSD}) \cite{liuSSDSingleShot2016} and \emph{Yolo v5} \cite{glenn_jocher_2022_6222936} solutions since they are suitable for low-resolution images and allows using a custom backbone without serious implications for the training process. In this study, we enforce the same architecture for feature extraction as in the AEs and CAMs approaches. However, unlike the previous techniques, such supervised mappings do not require the composition with the thresholding function $g_\tau$. Let $(x,b)\in \mathcal{F}$ be an IR image with its corresponding bounding box annotation. The learning process for both methods solves the optimization problem $\vartheta^* = \arg \min \mathcal{L}(x,b,\vartheta)$, by minimizing the cost of the model $\mathcal{L}$. Despite their differences in terms of representation, in both cases the loss function uses a supervised approach that measures the difference between the output $z=h_\vartheta(x)$ and the expected detections $b$.

\section{Experimental Methodology}

\subsection{Datasets}

In this study, two datasets were used to assess the IR person detection using models trained with different levels of annotation -- the public FIR-Image-Action~\cite{FIR-Image-Action-Dataset} dataset, and our Distech-IR dataset.

\noindent \textbf{1) FIR-Image-Action with bounding box annotations} 

The FIR-Image-Action~\cite{FIR-Image-Action-Dataset} dataset includes 110 annotated videos. We randomly selected 36 videos from this pool for the test and the others 74 for training and validation. Furthermore, training and validation sets were separated using a random selection of the frames (70\% and 30\%, respectively). All the approaches have been trained using the same data partition to ensure comparability.

To the best of our knowledge, there are no low-resolution IR datasets with bounding box annotations for person detection. Therefore, we annotated this dataset at bounding box level. The dataset was created by Haoyu Zhang of Visiongo Inc. for video-based action recognition. Such a dataset offers 126 videos with a total combined duration of approximately 7 hours. Since this study aims to evaluate the performance of different techniques for IR-based people localization, we only used the IR images provided by the authors for our experiments. Nevertheless, it is worth mentioning that two modalities are available within the dataset: RGB with a spatial resolution of $320 \times 240$ acquired at 24 FPS, and IR with a resolution of $32\times 24$ and sampled at 8 FPS. Although the RGB falls outside the scope of this work, we used them for obtaining bounding box annotations, as described later. As part of this work's contributions, we have publicly created and released the localization annotations for 110 videos out of the 126 for both RGB and IR modalities. Since there is redundancy within neighboring frames and our application does not require video processing, we further sampled the IR dataset obtaining the equivalent of 2 FPS videos.

We used a semi-automated approach to obtain bounding box annotations for the challenging low-resolution IR images in FIR-Image-Action. First, we create bounding box annotations by hand of a randomly selected subset of the RGB frames. We carefully curated these bounding boxes to reduce the impact of the misplacement when decreasing the resolution for the IR modality. Then, an \emph{SSD} detector \cite{liuSSDSingleShot2016}, $h_\vartheta$, was trained over the RGB annotated dataset, being used afterward to obtain pseudo-labels over the remaining unannotated partition of the RGB dataset. A new randomly selected subset is then curated and $h_\vartheta$ training is repeated but using a larger partition of the data. This process was repeated three times resulting in a fully annotated version of the RGB dataset.

\begin{figure}[ht]
\centering
\centering
    \begin{subfigure}{0.25\textwidth}
        \includegraphics[width=\textwidth]{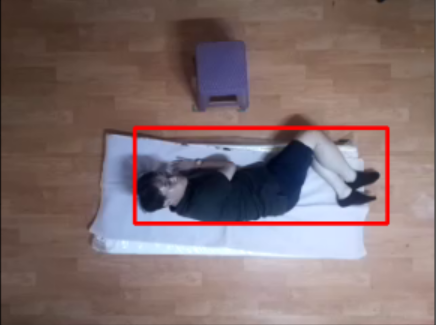}
        \label{fig:al1}
        \caption{}
    \end{subfigure}
    \begin{subfigure}{0.235\textwidth}
        \includegraphics[width=\textwidth]{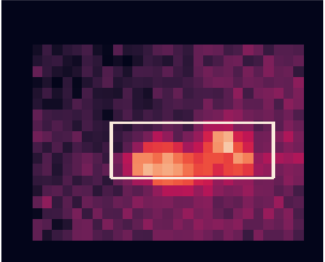}
        \label{fig:al2}
        \caption{}
    \end{subfigure}
    \begin{subfigure}{0.25\textwidth}
        \includegraphics[width=\textwidth]{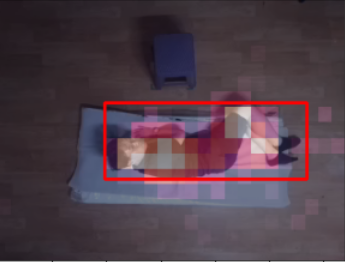}
        \label{fig:al3}
        \caption{}
    \end{subfigure}
\caption{Example of an RGB image with its ground truth (a), the corresponding IR image with the aligned bounding box (b), and an overlay of RGB and IR modalities (c) from the FIR-Image-Action dataset.}
\label{fig:alignment_images}
\end{figure}

Finally, bounding box annotations for the IR dataset are obtained by pairing images from both modalities, followed by a coordinate aligning procedure. Since the videos for IR and RGB were out of synchronization, the initial time shift was manually determined using an overlay visualization of both modalities (see \autoref{fig:alignment_images}c). Such a synchronization was performed individually for every video. The final IR localization annotations were obtained by doing a linear interpolation of the bounding box coordinates from the labeled RGB dataset. The parameters for the alignment were estimated using linear regression. An example of the obtained bounding box annotation for both RGB and IR modalities can be observed in \autoref{fig:alignment_images}a and b, respectively.

{\noindent \textbf{2) Distech-IR}}

\begin{figure}[ht!]
\centering
    \begin{subfigure}{0.22\textwidth}
        \includegraphics[width=\textwidth]{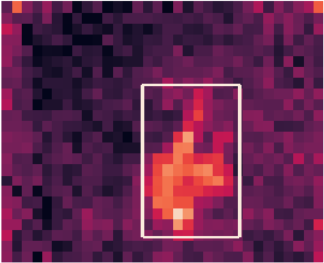}
        \label{fig:FIR}
        \caption{}
    \end{subfigure}
    \begin{subfigure}{0.22\textwidth}
        \includegraphics[width=\textwidth]{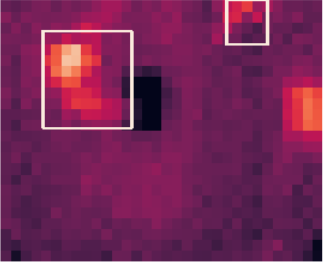}
        \label{fig:distech1}
        \caption{}
    \end{subfigure}
    \begin{subfigure}{0.22\textwidth}
        \includegraphics[width=\textwidth]{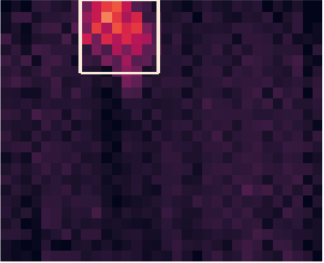}
        \label{fig:distech2}
        \caption{}
    \end{subfigure}
    \begin{subfigure}{0.22\textwidth}
        \includegraphics[width=\textwidth]{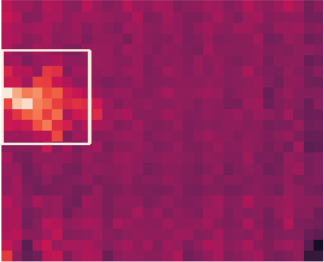}
        \label{fig:distech3}
        \caption{}
    \end{subfigure}
\caption{Examples of IR images with their corresponding ground truth for FIR-Image-Action (a) and Distech-IR (b)-(d) datasets.}
\label{fig:datasets}
\end{figure}

The second dataset, named hereafter Distech-IR, followed the same separation proportions containing 1500 images for training, 500 for validation, and 800 for testing. Such a dataset, similar to FIR-Image-Action, contains two modalities of images (RGB and IR) with their corresponding bounding box level annotations provided by Distech Controls Inc. The dataset reflects the increasing interest by the industry for privacy preserving-based solutions for person localization and constitute an actual use case for this task. The Distech-IR dataset also proved to reflect better real-world scenarios since it is composed of seven rooms with different levels of difficulties, i.e., heat radiating appliances, sun-facing windows, and more than one person per room. For simulating deployment, we used rooms not seen before during training for the test set. \autoref{fig:datasets} shows some examples of images from both datasets.

\subsection{Implementation details}

We used normalization by 50$^{\circ}$ to ensure small scale within the input map. Adam optimizer \cite{kingma2014adam} was employed with an initial learning rate of $10^{-4}$ and decay of 0.2 with a patience of ten epochs. Additionally, a 15 epoch patience early-stopping was implemented. Then, the best model according to the validation loss was selected for each case. The time calculations were evaluated on an Intel Xeon CPU at 2.3 Ghz, however the training of the models was done on an NVIDIA Tesla P100 GPU. Each experiment was performed 3 times with different seeds. The validation protocols are presented with each dataset in section 4.1.

\subsection{Performance metrics}

In this study, we use the optimal Localization Recall Precision (oLRP) \cite{oksuzOneMetricMeasure2021} in order to characterize the ability of each method to detect the presence of people and locate them. This metric allows us to evaluate with the same measurement methods that provide bounding box detections without a score associated (such as AEs and CAMs) and methods with a detection score (as SSD and Yolo v5 detectors). Additionally, as stated by the authors, it reflects the localization quality more accurately than other measurements, providing separate measures for the different errors that a detection method can commit. The metric takes values between 0 and 100, with lower values being better. As part of the metric computation, we also calculate the localization (oLRP$_{loc}$), False Positive (oLRP$_{FP}$), and False Negative (oLRP$_{FN}$) components which provide more insights of methods behavior.
Finally, execution time on the same hardware has been computed to obtain an approximation of the time complexity.

\begin{table}[!t]
\begin{center}
\caption{Performance of detection methods on the FIR-Image-Action dataset. All metrics are calculated with an IoU of 0.5.}
\label{table:FIR-Image-Action-Dataset}
\begin{tabular}{|l|c|c|c|c|c|}
\hline
     \textbf{Model} & \textbf{oLRP}$\downarrow$ & \textbf{oLRP}$_{loc}\downarrow$ & \textbf{oLRP}$_{FP}\downarrow$ & \textbf{oLRP}$_{FN}\downarrow$ & \textbf{Time(ms)}$\downarrow$\\
\hline \hline
    Threshold & $86.5 \pm 0.1$ & 32.3 & 45.3 & 44.2 & \ \ 0.4\\ 
    Otsu's Threshold & $83.5 \pm 0.1$ & 31.6 & 45.7 & 27.7 & \ \ 0.7\\
    dVAE & $74.7 \pm 1.3$ & 31.2 & 26.6 & 24.5 & 13.0\\
    dAE & $77.4 \pm 1.3$ & 30.4 & 31.2 & 29.1 & 12.3\\
    CAM & $85.1 \pm 1.1$ & 34.3 & 41.2 & 29.0 & 11.4\\
    GradCAM & $85.5 \pm 3.2$ & 34.5 & 43.0 & 32.1 & 24.3\\
    LayerCAM & $84.8 \pm 2.2$ & 34.9 & 37.2 & 33.1 & 25.6\\
    SSD & $63.8 \pm 2.7$ & \textbf{25.3} & 12.6 & 18.6 & 46.6\\
    Yolo v5 & $\textbf{56.9} \pm 1.8$ & 25.5 & \textbf{6.3} & \textbf{6.2} & 45.9\\
\hline
\end{tabular}
\end{center}
\end{table}
\setlength{\tabcolsep}{3pt}

\begin{table}[!b]
\begin{center}
\caption{Performance of detection methods on the Distech-IR dataset. All metrics are calculated with an IoU of 0.5.}
\label{table:Distech-Low-Resolution-IR-v9}
\begin{tabular}{|l|c|c|c|c|c|c|}
\hline
     \textbf{Model} & \textbf{oLRP}$\downarrow$ & \textbf{oLRP}$_{loc}\downarrow$ & \textbf{oLRP}$_{FP}\downarrow$ & \textbf{oLRP}$_{FN}\downarrow$ & \textbf{Time(ms)}$\downarrow$\\
\hline \hline
	Threshold & $93.6 \pm 2.4$ & 37.1 & 72.7 & 54.1 & \ \ 0.4\\
	Otsu's Threshold & $95.5 \pm 1.2$ & 34.2 & 83.3 & 50.0 & \ \ 0.7\\
    dVAE & $83.3 \pm 8.9$ & 33.2 & 32.7 & 40.6 & 13.0\\
    dAE & $82.7 \pm 9.0$ & 32.4 & 33.7 & 40.0 & 12.3\\
    CAM & $93.1 \pm 1.9$ & 37.6 & 59.7 & 52.3 & 11.4\\
    GradCAM & $91.6 \pm 2.3$ & 37.5 & 50.3 & 48.8 & 24.3\\
    LayerCAM & $91.1 \pm 2.5$ & 37.7 & 45.5 & 50.3 & 25.6\\
    SSD & $82.0 \pm 7.2$ & 31.1 & \textbf{26.3} & 44.7 & 46.6\\
    Yolo v5 & $\textbf{80.2} \pm 7.7$ & \textbf{30.2} & 31.4 & \textbf{37.4} & 45.9\\
\hline
\end{tabular}
\end{center}
\end{table}
\setlength{\tabcolsep}{3pt}

\section{Results and Discussion}

Tables \ref{table:FIR-Image-Action-Dataset} and \ref{table:Distech-Low-Resolution-IR-v9} summarize the obtained results for FIR-Image-Action and Distech-IR datasets, respectively. As expected, the fully supervised approaches obtained the best performance for most metrics and comparable results to the other approaches regarding False Negatives.
The methods proved useful for locating people in diverse scenarios, even under low-resolution settings. However, they take longer to execute than the second-best performed techniques, which is an important downside to take into account for measuring real-time occupancy levels in intelligent buildings. In particular, the \emph{dAE} approach was 3.7 times faster than \emph{Yolo v5} and 3.8 times faster than \emph{SSD}.

We can refer to the AE-based anomaly detection approaches as second-placed strategies. Both \emph{dAE} and \emph{dVAE} showed similar performance in terms of LRP and efficiency between them. Furthermore, the techniques obtained localization performance comparable to the fully-supervised approaches, especially in more complex situations like the Distech-IR dataset. This result is remarkable considering that localization supervision is not used during AE training, and only empty room images were used (10\% of the data in the FIR-Image-Action and 30\% in Distech-IR). The methods also showed acceptable execution times for the application.

\begin{figure}[!htb]
\centering
\setlength{\tabcolsep}{1pt}
\begin{tabular}{cccc}
 \includegraphics[width=.257\textwidth]{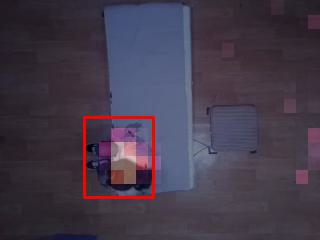} &
 \includegraphics[width=.237\textwidth]{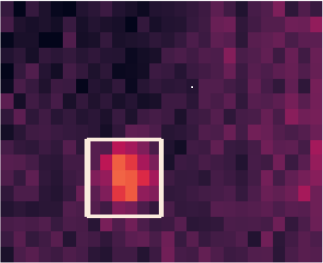} &
 \includegraphics[width=.237\textwidth]{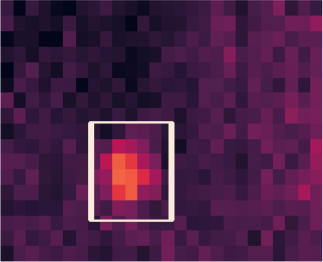} &
 \includegraphics[width=.237\textwidth]{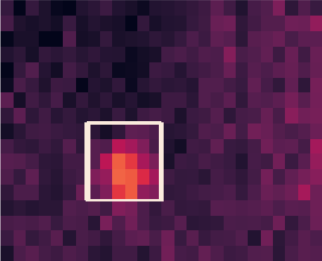}\\
 \includegraphics[width=.257\textwidth]{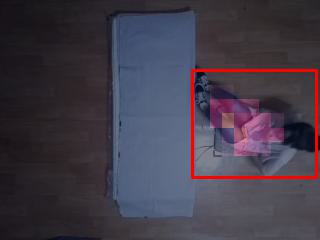} &
 \includegraphics[width=.237\textwidth]{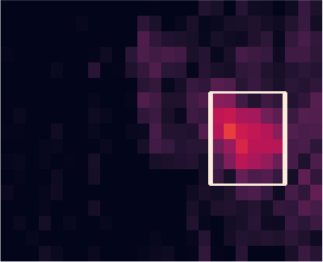} &
 \includegraphics[width=.237\textwidth]{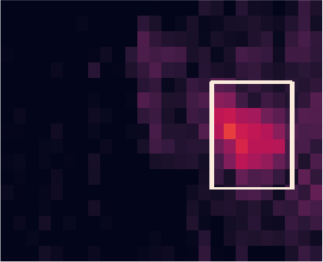} &
 \includegraphics[width=.237\textwidth]{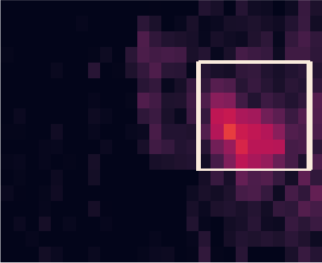}\\
 \includegraphics[width=.257\textwidth]{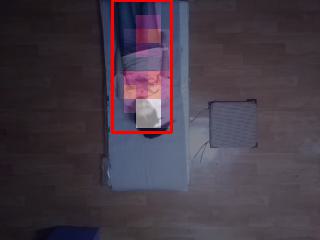} &
 \includegraphics[width=.237\textwidth]{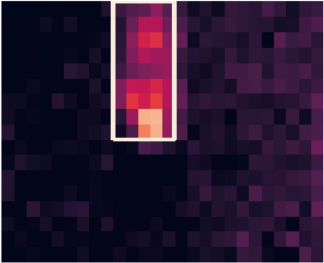} &
 \includegraphics[width=.237\textwidth]{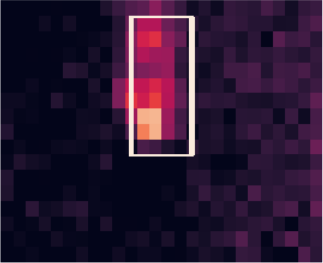} &
 \includegraphics[width=.237\textwidth]{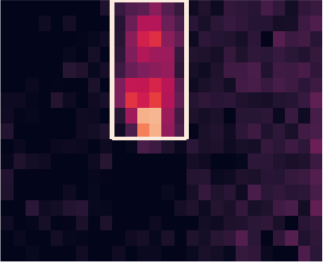}\\
 \includegraphics[width=.257\textwidth]{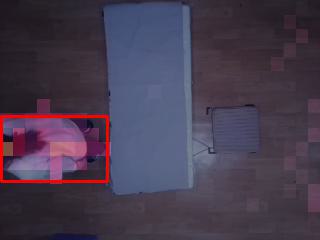} &
 \includegraphics[width=.237\textwidth]{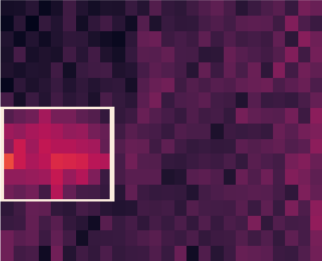} &
 \includegraphics[width=.237\textwidth]{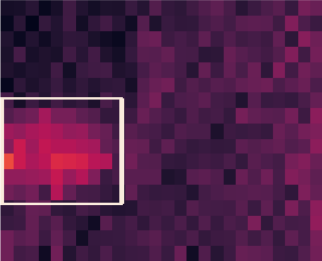} &
 \includegraphics[width=.237\textwidth]{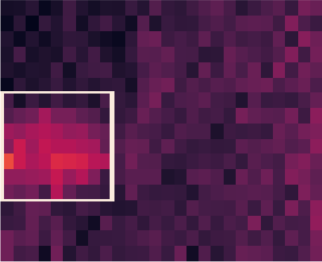}\\
 \includegraphics[width=.257\textwidth]{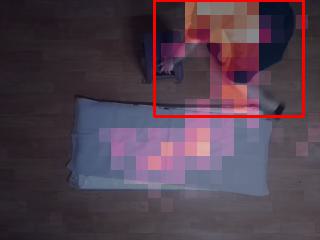} &
 \includegraphics[width=.237\textwidth]{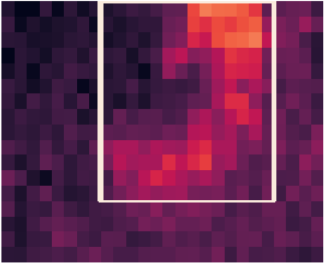} &
 \includegraphics[width=.237\textwidth]{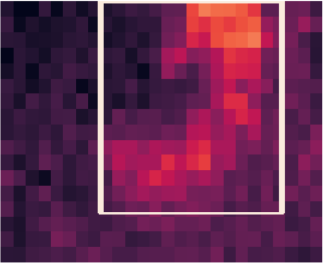} &
 \includegraphics[width=.237\textwidth]{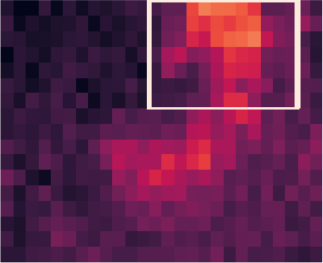}\\
 \includegraphics[width=.257\textwidth]{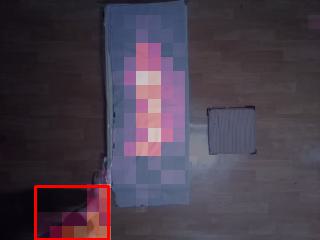} &
 \includegraphics[width=.237\textwidth]{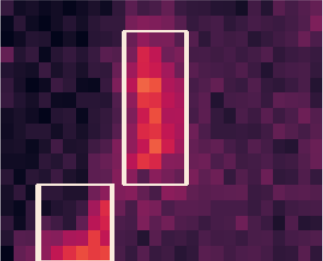} &
 \includegraphics[width=.237\textwidth]{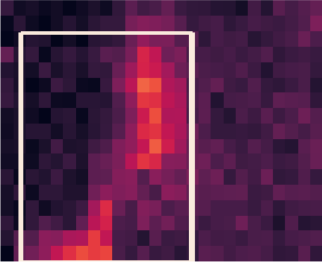} &
 \includegraphics[width=.237\textwidth]{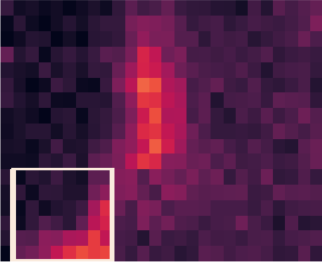}\\
 \includegraphics[width=.257\textwidth]{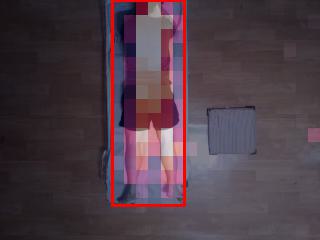} &
 \includegraphics[width=.237\textwidth]{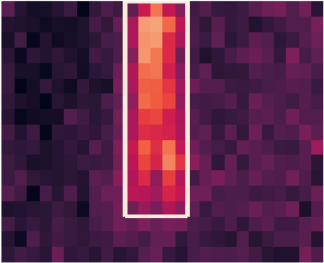} &
 \includegraphics[width=.237\textwidth]{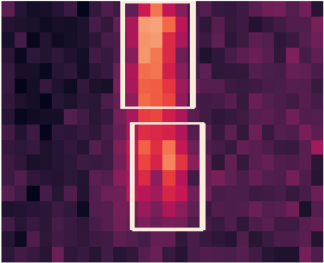} &
 \includegraphics[width=.237\textwidth]{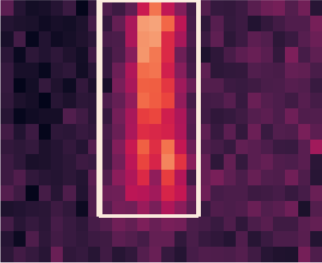}\\
 \includegraphics[width=.257\textwidth]{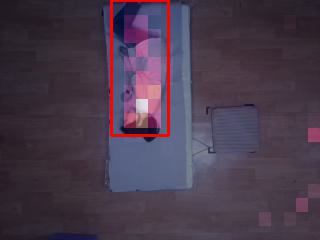} &
 \includegraphics[width=.237\textwidth]{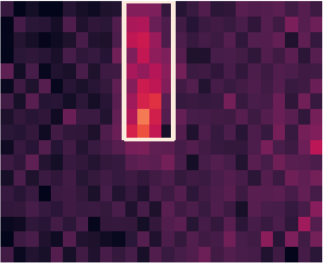} &
 \includegraphics[width=.237\textwidth]{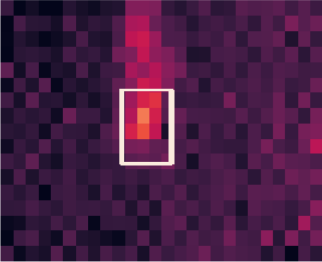} &
 \includegraphics[width=.237\textwidth]{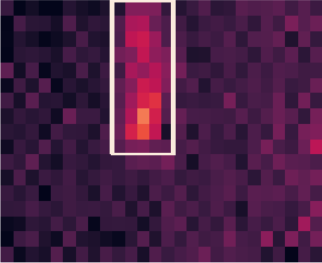}\\
 (a) & (b) & (c) & (d)\\
\end{tabular}
\caption{Examples of low-resolution IR people detection results. Overlay of RGB
and IR modalities with their corresponding ground truth (a), along with bounding box predictions of \emph{dVAE} (b), \emph{gradCAM} (c), and \emph{Yolo v5} (d).}
\label{detection_examples}
\end{figure}

CAM methods provide a lower level of performance than other methods, despite having access to class-label annotations for training. Indeed, these methods are known to activate strongly for discriminant regions of an input image (since the backbone CNN is trained to discriminate classes), and be affected by complex image backgrounds~\cite{belharbi2021f}. These two factors affect its ability to define precise contours around a person. 

As can be seen in the tables, \emph{Otsu's Threshold} provides good person localization. However, it assumes a multi-modal intensity distribution for finding the threshold, which leads to false person localization in empty rooms. This effect can be observed by the high values of oLRP$_{FP}$. The rest of the approaches showed comparable results to this last one but were still far from the fully-supervised process. \autoref{detection_examples} shows some examples of the obtained result over the FIR-Image-Action for \emph{dVAE}, \emph{gradCAM}, and \emph{Yolo v5} methods.

As expected, real scenarios like those depicted in Distech-IR proved harder to generalize. A decrease in the performance was observed in all levels of supervision with a significant drop of 18.2\% oLRP for \emph{SSD} and 23.3\% for \emph{Yolo v5}. A smaller decrease was observed for AEs obtaining even closer results to the one from supervised approaches. The primary issue in this dataset was the large number of False Negative which almost doubled the FN obtained for FIR-Image-Action.

\section{Conclusions}

In this work, we presented a study comprising different methods with increasing levels of supervision for privacy-preserving person localization. Our experimental results over two low-resolution top-view IR datasets showed that reduced image-level supervision is enough for achieving results almost comparable to a fully-supervised detectors. Specifically, AE-based approaches proved to perform similarly to Yolo v5 in real-world scenarios by only using images of empty rooms for training and with 3.7 times less execution time. Such a result is significant for reducing annotation costs and improving the scalability of intelligent building applications. Additionally, we detailed the process for producing bounding box annotations for low-resolution IR images and provided the localization for the publicly available dataset FIR-Image-Action.

\subsubsection{Acknowledgements:} This work was supported by Distech Controls Inc., and the Natural Sciences and Engineering Research Council of Canada (RGPIN-2018-04825).

%
%
\clearpage
\bibliographystyle{splncs04}
\bibliography{Bibliography.bib}
\end{document}